\documentclass{article}
\usepackage[table]{xcolor}
\usepackage{multirow}
\usepackage{array}
\usepackage{amsmath}
\usepackage{graphicx} % For including images
\usepackage{booktabs} % For better-looking tables
\usepackage{multirow} % For multi-row cells in the table
\usepackage{array}    % For controlling column width
\usepackage{amsmath}  % For mathematical symbols (e.g., x_n, t_n)
\usepackage{xcolor}   % For cell coloring (if desired)
\usepackage{lipsum}   % For generating placeholder text, remove in your actual paper
\usepackage{amssymb}
\usepackage{booktabs}
\usepackage{array}
\usepackage{geometry}
\usepackage{amsthm}  % 必需宏包

\usepackage{fontspec}
\usepackage{siunitx}
\usepackage{authblk}
\usepackage{makecell} % 用于处理多行表头
\usepackage{lineno}

\usepackage{xcolor}
\usepackage{listings}
\usepackage[skip=0.5\baselineskip]{caption}

\definecolor{codegreen}{rgb}{0,0.6,0}
\definecolor{codegray}{rgb}{0.5,0.5,0.5}
\definecolor{codepurple}{rgb}{0.58,0,0.82}
\definecolor{backcolour}{rgb}{0.95,0.95,0.92}

\lstdefinestyle{mystyle}{
    backgroundcolor=\color{backcolour},   
    commentstyle=\color{codegreen},
    keywordstyle=\color{magenta},
    numberstyle=\tiny\color{codegray},
    stringstyle=\color{codepurple},
    basicstyle=\ttfamily\footnotesize,
    breakatwhitespace=false,         
    breaklines=true,                 
    captionpos=b,                    
    keepspaces=true,                 
    numbers=left,                    
    numbersep=5pt,                  
    showspaces=false,                
    showstringspaces=false,
    showtabs=false,                  
    tabsize=2,
    frame=single,
    rulecolor=\color{gray}
}

\definecolor{lightblue}{RGB}{230, 240, 255}
\definecolor{lightgreen}{RGB}{230, 255, 240}

\title{Beyond Algorithm Evolution: An LLM-Driven Framework for the Co-Evolution of Swarm Intelligence Optimization Algorithms and Prompts}
\author[1]{Shipeng Cen}
\author[2]{Ying Tan}
\affil[1,2]{ School of Intelligence Science and Technology, Peking University, Beijing 100871, China}
\date{} % The date is usually set by the publication

\begin{document}

\maketitle
\begin{abstract}
The field of automated algorithm design has been advanced by frameworks such as EoH, FunSearch, and Reevo. Yet, their focus on algorithm evolution alone, neglecting the prompts that guide them, limits their effectiveness with LLMs—especially in complex, uncertain environments where they nonetheless implicitly rely on strategies from swarm intelligence optimization algorithms. Recognizing this, we argue that swarm intelligence optimization provides a more generalized and principled foundation for automated design. Consequently, this paper proposes a novel framework for the collaborative evolution of both swarm intelligence algorithms and guiding prompts using a single LLM. To enhance interpretability, we also propose a simple yet efficient evaluation method for prompt templates. The framework was rigorously evaluated on a range of NP problems, where it demonstrated superior performance compared to several state-of-the-art automated design approaches. Experiments with various LLMs (e.g., GPT-4o-mini, Qwen3-32B, GPT-5) reveal significantly divergent evolutionary trajectories in the generated prompts, further underscoring the necessity of a structured co-evolution framework. Importantly, our approach maintains leading performance across different models, demonstrating reduced reliance on the most powerful LLMs and enabling more cost-effective deployments. Ablation studies and in-depth analysis of the evolved prompts confirm that collaborative evolution is essential for achieving optimal performance. Our work establishes a new paradigm for swarm intelligence optimization algorithms, underscoring the indispensable role of prompt evolution.
\end{abstract}

\section{Introduction}
Optimization problems are ubiquitous across various facets of human society and are deeply intertwined with production and daily life. Yet, many real-world complex optimization problems are often NP, NP-hard or NP-complete\cite{goldreich2010p} in nature, for which no perfect algorithms running within polynomial time are currently available, and their search spaces are typically exponential in size. As research into the properties of these problems progresses, investigators have begun employing optimization algorithms such as swarm intelligence algorithms\cite{bansal2019evolutionary} and evolutionary algorithms\cite{back1993overview} to explore the solution space. However, on one hand, the design of various rules within these algorithms heavily relies on expert knowledge; on the other hand, the continuous emergence of new optimization methods has made the selection of high-performing algorithms a challenging issue for researchers. These two factors have posed certain obstacles to the advancement of the field. Furthermore, when confronted with complex optimization problems, researchers tend to first turn to classical and efficient algorithms like PSO\cite{wang2018particle}, DE\cite{das2010differential}, FWA\cite{tan2010fireworks}, and CMA-ES\cite{hansen2003reducing}, etc. Should these methods still yield unsatisfactory results, they often need to trial other algorithms—a process that can be highly inefficient.

With the rise of large language models(LLMs)\cite{zhao2023survey}, automatic algorithm design\cite{liu2024systematic} based on LLMs has attracted widespread scholarly attention and growing interest. One the one hand, LLMs acquire vast amounts of knowledge during training that far exceeds that of any individual, while demonstrating competence in problem understanding and algorithm design comparable to domain experts—thus providing a solid foundation for such applications. On the other hand, automated design frameworks leveraging LLMs significantly reduce the traditional heavy reliance on expert knowledge, enabling rapid, straightforward, and efficient application across diverse problem types. It can be said that the emergence of LLMs has injected new vitality into the field of optimization. Several notable works—such as EoH\cite{liu2024evolution}, Funsearch\cite{romera2024mathematical}, Reevo\cite{ye2024reevo} and AlphaEvolve\cite{novikov2025alphaevolve}—have already achieved performance surpassing traditional methods on many classic problem classes (e.g., TSP\cite{matai2010traveling}, CVRP\cite{toth2002vehicle}), while also setting numerous new state-of-the-art results. However, we observe that:

\begin{itemize}
    \item On certain challenging problem classes—where challenges may stem from constraint complexity or problem novelty—online constructive methods designed by these automated frameworks often underperform. As a result, they frequently resort to search strategies from various swarm intelligence algorithms or even directly invoke solvers for assistance. This suggests that relying entirely on online construction is impractical, and that search mechanisms from population-based optimization algorithms still play a critical role.
    \item While there has been growing work on prompt evolution—since different LLMs exhibit varied preferences for prompt phrasing—most mainstream automated algorithm design frameworks still overlook the impact of prompt design. Although they may generate different code in each interaction, other parts of the prompt template remain entirely human-specified, which ultimately limits the full exploitation of the model’s capabilities.
\end{itemize}

As for swarm intelligence optimization algorithms, existing research has begun to explore the integration of large language models. For example, some studies have utilized LLMs to design original operators in the fireworks algorithm and introduced novel operators to balance competition and cooperation within the population, with validation conducted on engineering design tasks. Other work has incorporated multimodal large language models in visually rich scenarios, thereby providing more comprehensive information to support MLLM-based decision-making. These efforts have also expanded the application scope of swarm intelligence optimization operators, with the overall architecture achieving outstanding performance on tasks such as the TSP and Electronic Design Automation (EDA)\cite{lin2019dreamplace,liao2022dreamplace}. However, similar to earlier frameworks such as EoH\cite{liu2024evolution}, these approaches still overlook the impact of prompt templates, which limits the full exploitation of the capabilities of LLMs.

Based on the current state of research, we propose a novel framework, which realizes the collaborative evolution of swarm intelligence algorithms and prompts based on large language models. This framework is designed to address the previously overlooked aspect of prompt evolution in automated algorithm design. Both the evolution of the algorithms themselves and the evolution of the prompt templates are driven by the same large language model. Furthermore, we introduce a simple yet effective evaluation method for prompt templates, which enhances the interpretability of the prompt evolution process. For the experimental validation, we tested our framework on various NP problems with LLMs like gpt-4o-mini\cite{hurst2024gpt}, Qwen3-32B\cite{yang2025qwen3} and GPT-5\cite{gpt5}, and conducted comparisons with several mainstream approaches, the results of which demonstrate the superiority of our proposed framework. Additionally, ablation studies and an analysis of the generated prompt templates confirm the necessity of integrating collaborative prompt evolution.

\section{Related Work}
\subsection{Swarm Intelligence Algorithms}
Swarm intelligence optimization algorithms represent a class of multi-population optimization algorithms whose performance depends not only on the exploration and exploitation capabilities of individual agents but also on the interactions and cooperation among them. These algorithms offer advantages such as gradient-independent operation, ease of parallelization, and simple rules. Historically, the design of swarm intelligence algorithms has drawn upon various "inspirations," which may originate from the behaviors of certain organisms (e.g., ACO\cite{blum2005ant}, PSO\cite{wang2018particle}, GWO\cite{mirjalili2014grey}), natural phenomena (e.g., FWA\cite{tan2010fireworks}), or even human activities (e.g., BSO\cite{shi2015optimization}), among other sources.

A well-designed swarm intelligence optimization algorithm requires carefully crafted operators(rules) and rigorous benchmark testing, ideally supported by convergence theory—though this aspect remains relatively underdeveloped in the field. Currently, swarm intelligence optimization algorithms have found extensive applications across various domains, including traditional problems such as TSP and CVRP, neural architecture\cite{byla2019deepswarm} and hyperparameter search\cite{lorenzo2017particle}, and even low-rank fine-tuning of large language models under very limited GPU memory constraints\cite{jin2024derivative}.

\begin{figure}
    \centering
    \includegraphics[width=1\linewidth]{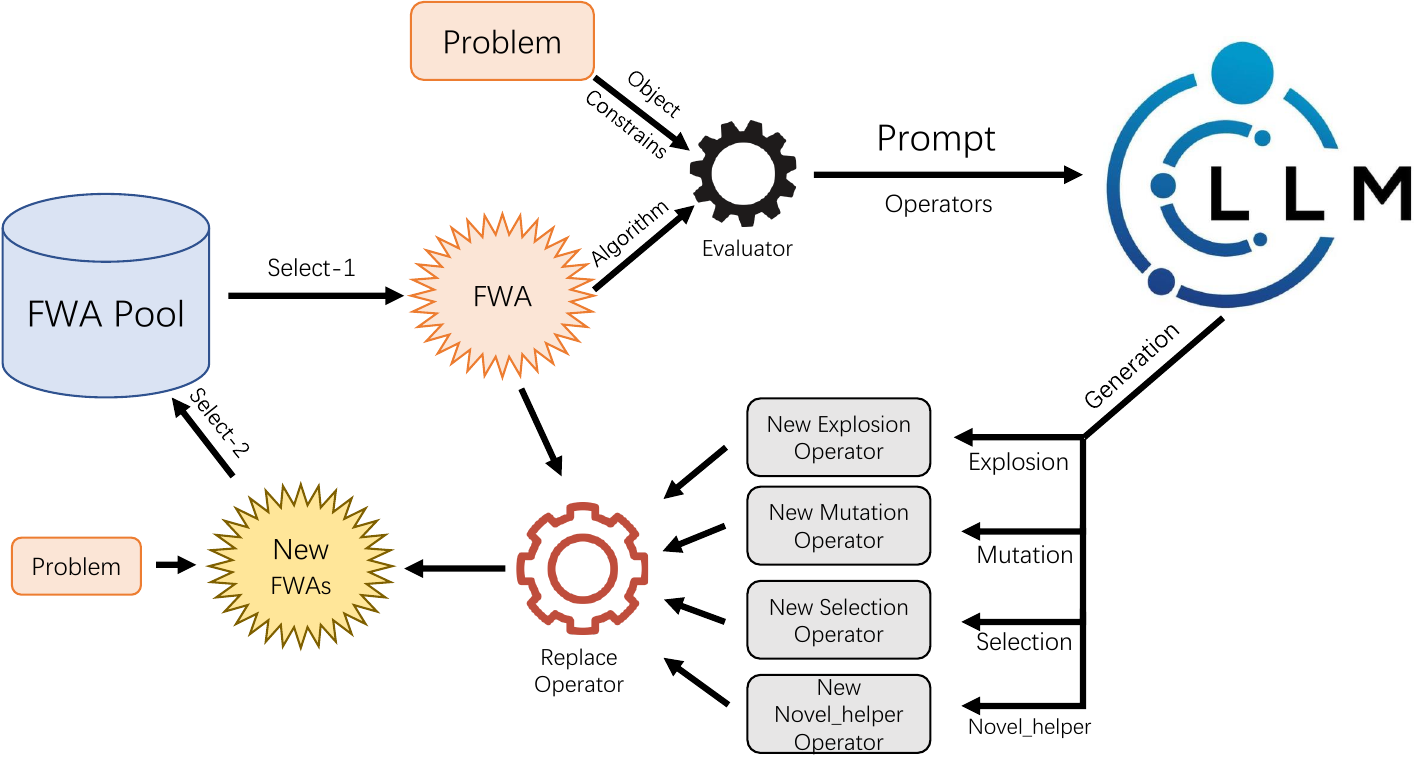}
    \caption{LLM-driven pipeline for Fireworks Algorithm (FWA) evolution from \cite{cen2025llm}.}
    \label{fig:llm+fwa}
\end{figure}

The design of effective operators is often task-specific and poses a significant challenge for human experts. In response, some researchers have begun to integrate large language models with FWA\cite{cen2025llm}, developing an automated evolutionary framework to generate high-performance operators tailored to specific tasks, as shown in Figure \ref{fig:llm+fwa}.

Why choose the Fireworks Algorithm? On the one hand, compared to other swarm intelligence algorithms, FWA is more of a framework-oriented approach, featuring multiple operators—such as explosion, mutation, and selection operators—thus offering a rich design space for operator customization. On the other hand, both the conventional FWA and its LLM-enhanced variants have demonstrated strong performance across a variety of tasks. This paper will use FWA as a case study to construct a novel LLM-based evolutionary framework different from before.

\subsection{LLM-driven Automatic Algorithm Design}
In recent years, research on automatic algorithm design leveraging Large Language Models has yielded remarkable breakthroughs, showcasing the significant potential of this paradigm to revolutionize traditional algorithm design processes. Notable representative works highlight these advancements. For instance, FunSearch\cite{romera2024mathematical} combines the code generation capability of LLMs with an automated evaluator, enabling the discovery of solutions surpassing known human-derived results in mathematical and combinatorial optimization problems. EoH\cite{liu2024evolution} introduces a concept of "idea-code" co-evolution, utilizing diverse prompting strategies to drive LLMs in automatically generating high-performance heuristic algorithms, demonstrating exceptional design efficiency in tasks like online bin packing and the Traveling Salesman Problem. ReEvo\cite{ye2024reevo} incorporates dual-level (short-term and long-term) reflection mechanisms, enabling LLMs to continuously integrate experiences and refine heuristics by learning from comparative performance analysis. AlphaEvolve\cite{novikov2025alphaevolve} is a novel AI-powered coding agent developed by DeepMind, designed for scientific and algorithmic discovery. This system has demonstrated remarkable versatility by achieving breakthroughs in both theoretical domains, such as discovering new matrix multiplication algorithms and improving solutions to long-standing mathematical problems like the kissing number problem, as well as in practical applications, including optimizing data center scheduling and hardware design.

The common core advantages of these works include an exceptionally high level of automation, significantly reducing reliance on human expertise and manual coding; strong potential for algorithmic innovation, as LLMs can combine or generate novel algorithm components beyond conventional design thinking; and notable generality and flexibility, enabling adaptation to various algorithm design tasks.

As for swarm intelligence optimization algorithms, \textbf{we believe their deep integration with LLMs is not only an inevitable trend in technological development but also highly necessary.} The core strength of swarm intelligence algorithms lies in their ability to effectively explore complex solution spaces through interaction and cooperation among population individuals. However, their performance heavily depends on the design of various operators (e.g., position update, pheromone update, explosion, mutation), which often rely on fixed, human-crafted rules. This can limit their adaptability when facing novel, dynamic complex optimization tasks. The introduction of LLMs provides a key pathway to address this core bottleneck. LLMs can act as efficient "meta-designers," automatically designing or optimizing core operators for specific swarm intelligence algorithms. For example, exploratory work has begun combining LLMs with FWA, utilizing LLMs to automatically design more effective explosion, mutation, and selection operators for FWA, thereby significantly expanding the design space and potential of the FWA framework, as shown in Figure \ref{fig:llm+fwa}. Furthermore, the natural language interface of LLMs considerably lowers the barrier to algorithm design, allowing non-expert users to customize dedicated swarm intelligence optimization strategies through instructions, thereby promoting the democratization of this technology.

\subsection{Prompt Engineering}
In recent years, prompt engineering has evolved from a manual, experience-dependent craft into a systematic and automated engineering discipline. Among these advancements, automatic prompt evolution has emerged as a key research focus, aiming to leverage algorithms for dynamically generating and optimizing prompts, thereby enhancing the performance of large language models on specific tasks.

Researchers directly use LLM to generate and iterate prompts. For instance, Microsoft's PromptWizard\cite{agarwal2024promptwizard} framework implements fully automated discrete prompt optimization: it employs a feedback-driven process of criticism and synthesis to balance exploration and exploitation, iteratively refining prompt instructions and in-context examples to generate human-readable prompts tailored for specific tasks. This framework demonstrates excellent performance across 45 tasks, even with limited training data or when using smaller LLMs. Another relevant work is TextGrad\cite{yuksekgonul2025optimizing}, a framework in which researchers conceptualize "differentiation" as a metaphor: the feedback provided by LLMs is referred to as "text gradients." When optimizing prompts, TextGrad first evaluates the final prediction outcomes and then adjusts the initial prompts based on the feedback, thereby iteratively refining the entire process. Analogous to numerical gradient descent, Text gradient descent leverages optimization suggestions generated by language models to dynamically adjust system variables for performance enhancement.

Also, gradient of LLM can be used to further optimize prompts. The GradPromptOpt\cite{zhong2025gradpromptopt} method is an automatic gradient-based optimization framework. It leverages internal gradient information of the LLM to automatically diagnose and optimize weak components within a prompt, achieving a performance improvement of 31.75\% in fine-grained tasks like keyword extraction, far exceeding the efficiency of traditional manual tuning.

Prompts have demonstrated critical benefits for tasks; however, current automatic algorithm design frameworks primarily emphasize the algorithms themselves, overlooking the iterative refinement of prompts during the process.

\section{Methods}
\begin{figure}
    \centering
    \includegraphics[width=1\linewidth]{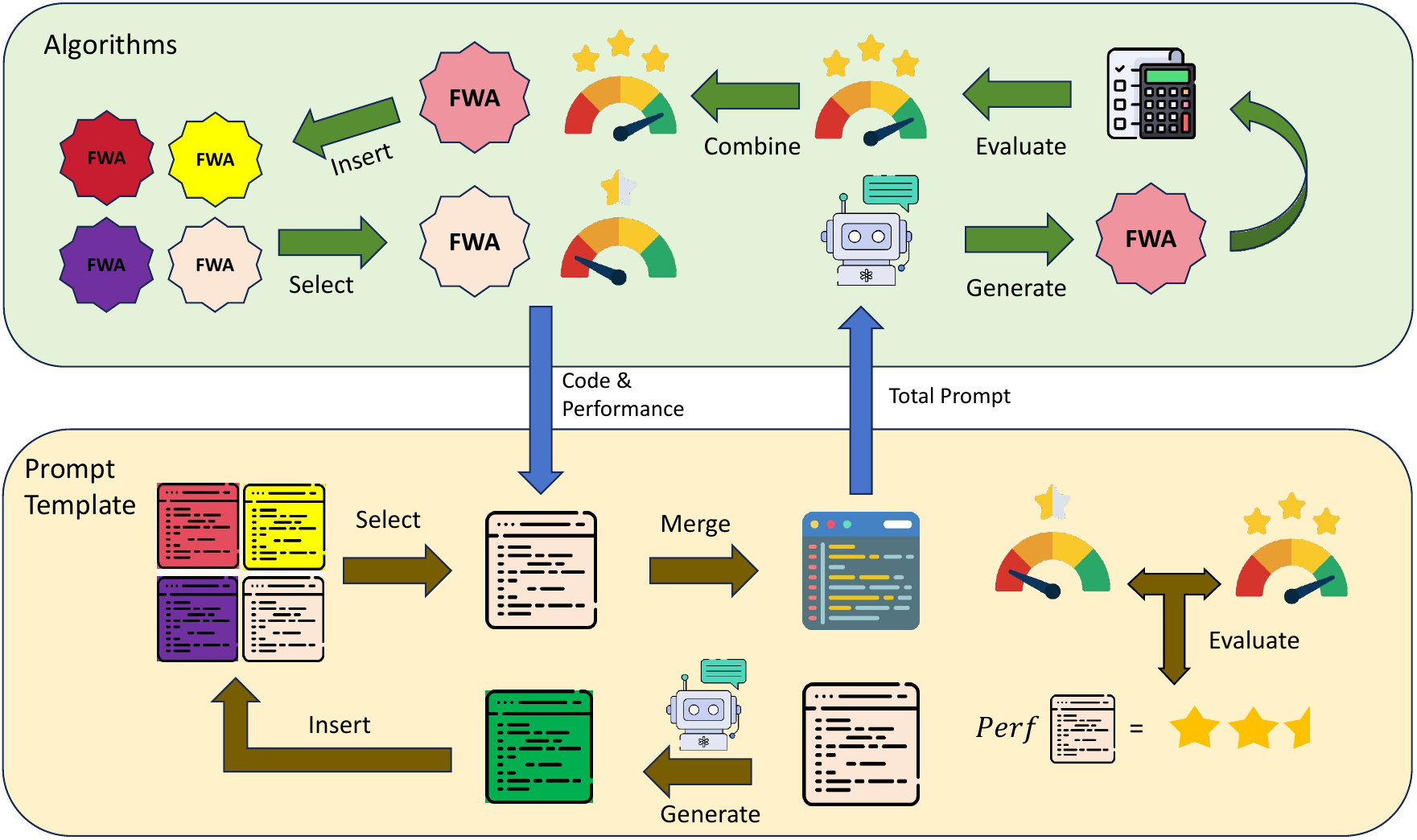}
    \caption{Swarm intelligence optimization algorithm and prompt template co-evolution framework based on large language model.Top subfigure: The iterative optimization process of a swarm intelligence optimization algorithm, illustrated here using the Fireworks Algorithm (FWA) as an example. Bottom subfigure: The automated evolution process of prompt templates, whose generation and updates are dependent on a LLM. The two processes are designed to interact with each other. It should be noted that there can be multiple types of prompt templates, meaning the framework may maintain multiple distinct prompt template pools.}
    \label{fig:co-evolution}
\end{figure}

Traditional automated algorithm design frameworks overlook the importance of interactive prompts, and the operational templates for interacting with large models remain static. To address this limitation, we devise a novel approach termed "algorithm-prompt co-evolution", with the entire process illustrated in Figure 1.

In essence, our method can further leverage the potential of large language models without requiring additional training. This is achieved by quantifying the contribution of individual prompt templates rather than the entire prompt used by PromptWizard during the algorithm evolution process, and dynamically adjusting these templates for various interactive operations (e.g., mutation, crossover). Crucially, adopting the entire prompt as the unit of design would significantly increase computational overhead and undermine generality, as the combinatorial explosion of possible prompt variations would render the optimization process intractable. In contrast, the template-level approach enables efficient and reusable prompt component optimization, enhancing both scalability and adaptability across diverse tasks.

\subsection{Algorithm Part}
We begin by discussing the evolution of population-based intelligence algorithms within the context of large language models, focusing specifically on the optimization algorithms themselves. This process comprises several key steps: algorithm selection, individual generation, task evaluation, and algorithm update.

First, we discuss algorithm selection. During the evolutionary process, it is necessary to choose candidate algorithms for evolution from the current algorithm set. A common approach is probability-based or greedy selection, where the probability can be derived from the actual evaluated values or performance metrics of the algorithms, as shown below, lower rank meaning better performance. Additionally, the number of algorithms required may vary depending on the genetic operator. For instance, a mutation operator typically selects only one algorithm, whereas a crossover operator may require two or more.

\[
p(A_i)= \frac{n + 1- rank(A_i)}{\sum_j rank(A_j)}, rank(A_i) \in \{1,2,3,...,n\} 
\]

\begin{figure}
    \centering
    \includegraphics[width=1\linewidth]{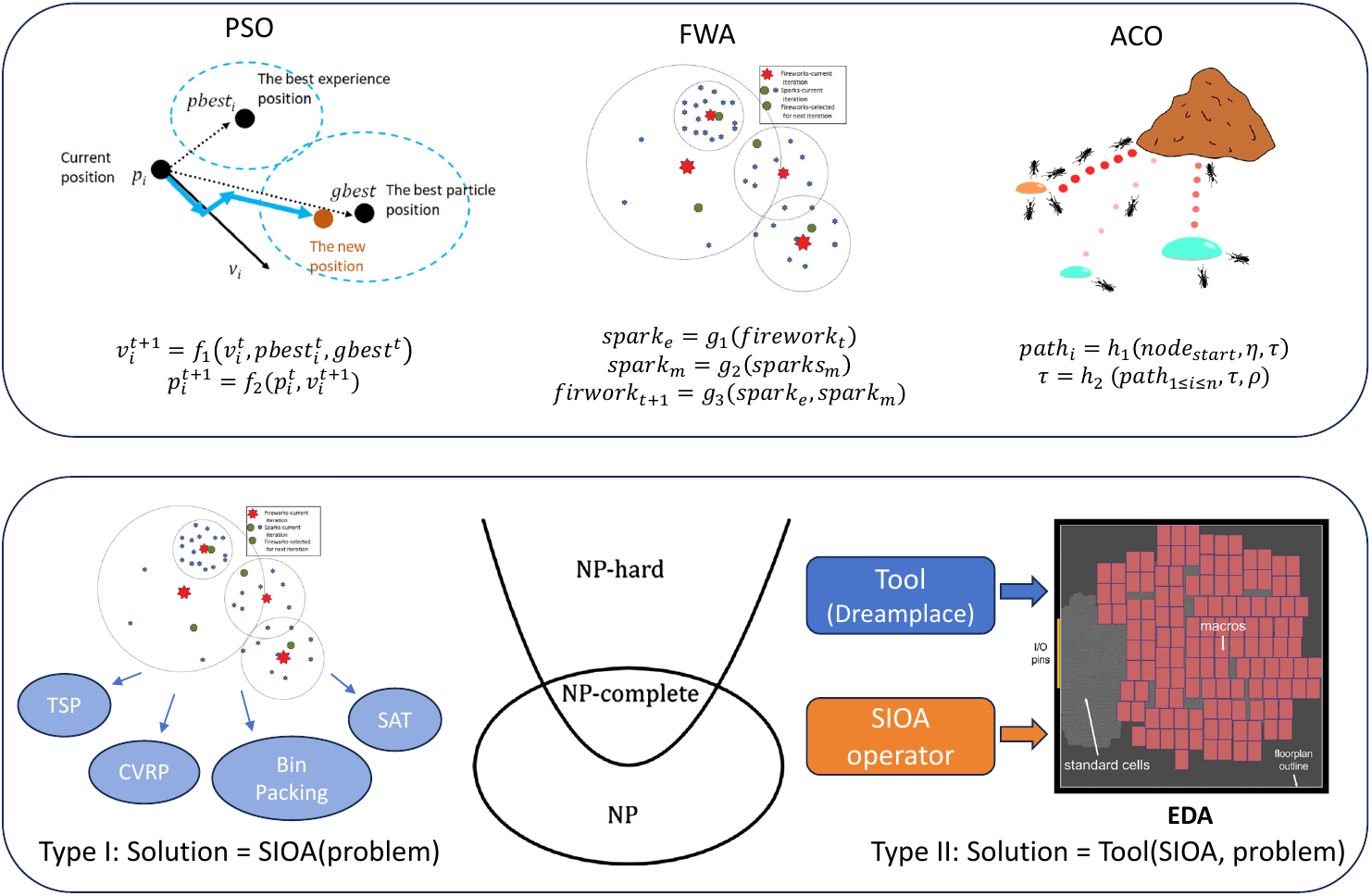}
    \caption{\textbf{Top subfigure}: The diagram includes three swarm intelligence algorithms—PSO, FWA, and ACO—along with their fundamental update rules. We use abstract functions such as f, g, and h to represent specific mathematical expressions, aiming to illustrate that: (1) these abstract functions can serve as design targets for the LLM, and can be evolved either individually or collectively; (2) these abstract functions essentially function as independent heuristic rules, which can be applied separately in certain contexts. \textbf{Bottom subfigure}: Some NP-hard optimization problems can be directly encoded using population-based intelligent optimization algorithms (Type I). However, some NP-hard problems are addressed using specialized tools that also incorporate many heuristics. In such cases, ideas from swarm intelligence algorithms can be leveraged to enhance the performance of these tools (Type II).}
    \label{fig:operator}
\end{figure}

Secondly, we proceed to introduce individual generation. For swarm intelligence algorithms and evolutionary algorithms, different methods may involve distinct procedures. For instance, FWA includes explosion, mutation, and selection operators; DE employs a differential update operator; while PSO involves position and velocity update operators. When working with a specific algorithm, it is essential to first clarify the object and granularity of generation. One may choose to update only a single operator in FWA, redesign all of its operators—though the latter entails more significant modifications—or adjust a single velocity update equation in PSO. Therefore, the term "key components" is more appropriate than "individual generation" in this context. A major advantage of utilizing large language models for generating such key components lies in their flexibility. For example, to address the competition and collaboration within populations in FWA, researchers have proposed a "novel helper" operator, which allows the LLM to freely implement code based on given prompts. Furthermore, traditional swarm intelligence optimization algorithms are typically derived from various conceptual inspirations—whether from biological systems or natural phenomena—and are manually crafted into code. These algorithms can generally only be utilized as complete optimizers, requiring extensive coding, selecting, testing, and other complex procedures. By leveraging large language models, however, these very concepts can serve as a design inspiration for specific operators tailored to difficult or complex tasks. This approach liberates swarm intelligence algorithms from their traditional role as full-scale optimizers, enabling their application at the operator level in tough problems like EDA, as shown in Fig.\ref{fig:operator}

The task evaluation component is relatively straightforward. We typically conduct multiple assessments of the generated algorithms, which can be accelerated through parallel implementation. Since population-based optimization algorithms require extensive search exploration, to prevent excessive complexity in the algorithm pool, we can employ time-constrained evaluation. This approach ensures that the final algorithms achieve an optimal balance between computational efficiency and solution quality.

In the algorithm update phase, the challenge we face is how to select among existing algorithms. Typically, a greedy strategy can be adopted to retain the top-performing algorithms. However, we must acknowledge that when two distinct algorithms A and B undergo iterative design improvements using large language models, they inherently explore different "algorithm spaces." Consequently, an algorithm with currently suboptimal performance might belong to an algorithm space that contains highly promising variants. Based on this insight, some studies have proposed Monte Carlo tree search-inspired approaches to balance immediate performance and potential future rewards.

\subsection{Prompt Part}
Previous studies have predominantly focused on the iterative generation of algorithms while overlooking the evolution of prompts. In practice, for a fixed LLM, different prompting strategies can lead to substantially different outcomes. Since each interaction prompt incorporates parameters such as the current individual's code and performance metrics, what we are effectively evolving is the prompt template. We aim for this prompt template—designed by the current LLM—to progressively adapt in a direction that maximizes compatibility with the LLM's characteristics, rather than relying on simplistic automated prompt engineering tools.

First, we need to identify which prompt templates require evolution. In this work, we designed two types of initial prompts: mutation and crossover. The former is intended to mutate a specific operator of the current population-based intelligent optimization algorithm, without prescribing the exact direction of mutation but only requiring a redesign. The latter aims to perform crossover blending on an operator from two distinct algorithms, without specifying the precise crossover method. The initial prompts were manually designed as follows:

\begin{lstlisting}[language=Python, caption={Mutation Prompt Generation Function}]
def get_mutation_prompts(problem_description, 
                         current_code, 
                         current_performance):
    introduction = (
        f"You are an expert in combinatorial optimization and "
        f"firework algorithms, now you are faced with a problem: "
        f"{problem_description}.\n"
    )
    current_fwa = (
        f"I will show you one firework algorithm for solving it:"
        f"{current_code}. Its performance is {current_performance}. "
        f"The higher performance, the better.\n" 
    )
    requirements = (
        "1. Choose exactly one function in 'explode', 'mutate' or 'select' "
        "and replace it with your new design\n"
        "2. Redesign only the chosen function, keeping input/output unchanged\n"
        "3. Return complete code in format: <code>xxx</code> without explanations"
    )
    
    return introduction + current_fwa + requirements
\end{lstlisting}

\begin{lstlisting}[language=Python, caption={Crossover Prompt Generation Function}]
def get_crossover_prompts(problem_description, 
                          current_codes, 
                          current_performances):
    introduction = (
        f"You are an expert in combinatorial optimization and "
        f"firework algorithms, now you are faced with a problem: "
        f"{problem_description}.\n"
    )
    current_fwa = (
        f"I will show you two firework algorithms for solving it.\n"
        f"First algorithm: {current_codes[0]}\n"
        f"Performance: {current_performances[0]}\n\n"
        f"Second algorithm: {current_codes[1]}\n"
        f"Performance: {current_performances[1]}\n"
        f"The higher performance, the better.\n" 
    )
    requirements = (
        "1. Perform crossover on 'explode', 'mutate' and 'select' functions "
        "using elements from both algorithms\n"
        "2. Maintain original input/output interfaces for all functions\n"
        "3. Return complete code in format: <code>xxx</code> without explanations"
    )
    
    return introduction + current_fwa + requirements
\end{lstlisting}

While the performance of an algorithm can be directly evaluated through task-specific assessments, there is no direct quantitative method for evaluating prompts. To address this, we posit that when a large language model is fixed, the quality of a prompt should be measured by the performance improvement it imparts to the algorithms it generates. As shown in the following equation, the performance of a prompt template can be measured by the performance difference between algorithms at consecutive time steps. It should be noted that this performance difference is actually dependent on the LLM. However, since the LLM remains frozen during application and the prompt template is intrinsically coupled with the LLM, we can disregard this dependency and attribute the performance gain entirely to the contribution of the prompt template. At the same time, it is well-known that improving a weak algorithm is relatively straightforward, whereas further enhancing an already strong algorithm is considerably more challenging. Therefore, the performance improvement achieved by a prompt template should be assigned a lower weight when applied to weak algorithms, a principle that is inherently consistent with our prompt template performance calculation method.

\[
Perf(Prompt_t)= E_{A_t\sim p(A)}[A_{t+1}(task)-A_{t}(task)|A_{t+1} = LLM(A_t, Prompt_{t})]
\]

In each round, we employ a selection mechanism similar to that used for algorithm selection, choosing prompt templates from the prompt template library based on performance rankings. The selected templates will incorporate the task description, code, and algorithm performance metrics to form complete prompts, which are then fed to the large language model for generating the next generation of algorithms. Periodically, we update the prompt template library using an interaction-based approach with the large language model, as illustrated below:

\begin{lstlisting}[language=Python, caption={Prompt Template Evolution Function}]
def getprompt4prompt(old_prompt_function):
    prompt4prompt = (
        f"You are an expert in prompt engineer. "
        f"I have a function for generating prompt below: {old_prompt_function}, "
        f"please help me to make it more powerful and appropriate for "
        f"automatic algorithm design.\n"
        "You can only modify the 'introduction', 'current_fwa', and "
        "'requirements' fields, not the input and output of the function, "
        "and make sure that this prompt allows the large language model to "
        "put the returned code in <code>xxx</code> for easy parsing.\n"
        f"Only return the code of the function for generating the prompt "
        f"in <prompt>xxx</prompt>. Do not explain anything"
    )
    return prompt4prompt
\end{lstlisting}

\section{Experiments}

In the experimental section, we note that while previous work has demonstrated the potential of LLM-based population intelligence algorithms on continuous problems (including black-box function test suites and low-dimensional constrained engineering design problems), these problems suffer from low dimensionality and insufficient relevance to real-world production scenarios, thus lacking persuasiveness. To address this limitation, we selected four distinct task categories from CO-Benchmark\cite{sun2025co} for comparison: aircraft landing problem, equitable partition problem, flow shop scheduling problem, and uncapacitated p-median problem. These problem instances vary in scale and feature diverse data distributions.

Our comparative baselines include state-of-the-art LLM-based automated algorithm design methods such as EoH, Funsearch, and Reevo, as well as traditional widely-applied population intelligence optimization algorithms like DE and PSO. To ensure fairness, we limited the maximum number of LLM-generated algorithms to 200 per experiment (preventing potential unfairness introduced by generating multiple algorithms in a single iteration) and run 5 times to get the best algorithm. 

Moreover, it is important to emphasize that, in all experiments, we \textbf{do not employ} penalty function methods to handle constraints. If a generated solution fails to satisfy the problem constraints, it is directly considered invalid. This approach further enhances the algorithm's capability and requirement for constraint handling. In the experiment, we selected the firework algorithm as the swarm intelligence optimization algorithm to be evolved.

All problem instances come with reference solutions. The performance of the algorithm is evaluated by comparing the solution obtained by the best algorithm generated during the process with the reference solution, using a ratio metric. Ideally, this ratio should be a real number within the interval $[0,1]$, though it may exceed 1 if a solution superior to the reference is found.

\[
\text{Performance} = 
\begin{cases}
\dfrac{\text{Best}}{\text{Reference}}, & \text{for maximization problem} \\\\
\dfrac{\text{Reference}}{\text{Best}}, & \text{for minimization problem}
\end{cases}
\]

\subsection{Aircraft landing problem}
This task addresses the aircraft landing scheduling problem, which involves assigning landing times and runway allocations to a set of aircraft under time window and separation constraints. The objective is to minimize the total penalty cost incurred from deviations between the actual landing times and the target times for each aircraft. Formally, given a set of aircraft $\mathcal{P}$  and a set of runways $\mathcal{R}$, the problem requires determining a landing time $t_p$ and a runway assignment $r_p \in \mathcal{R}$ for each aircraft $p \in \mathcal{P}$. Each aircraft $p$ is characterized by the following parameters: an appearance time $a_p$ (when the aircraft enters the system), an earliest landing time $e_p$, a target landing time $\tau_p$, a latest landing time $l_p$, a penalty rate $\alpha_p$ for early landing per unit time, and a penalty rate $\beta_p$ for late landing per unit time. The landing time $t_p$ must satisfy the constraint $e_p \leq t_p \leq l_p$. The cost for aircraft $p$ is computed as $\alpha_p \max(0, \tau_p - t_p) + \beta_p \max(0, t_p - \tau_p)$, representing linear penalties for earliness or lateness relative to $\tau_p$.

Separation constraints are imposed to ensure safe operations: if two aircraft $i$ and $j$ are assigned to the same runway, and if $t_i \leq t_j$, then the time gap must satisfy $t_j - t_i \geq s_{ij}$, where $s_{ij}$ is the separation time provided in the input matrix $s_{n \times n}$. If aircraft are assigned to different runways, no separation constraint applies between them, as the input matrix is defined specifically for same-runway assignments. The freeze time parameter is provided but does not influence scheduling decisions.

A solution is feasible only if all time window and separation constraints are satisfied. If any constraint is violated, the solution is deemed invalid and receives no score. The goal is to find a feasible schedule that minimizes the sum of penalty costs across all aircraft. Here we focus one the Aircraft landing instances where $\mathcal{|R|}=1$. In this task, with $\mathcal{|R|}$ growing, it is easier to design a solution that satisfies the constraints.

In this problem, the encoding scheme of our initial fireworks algorithm represents the takeoff sequence of different aircraft, transforming the time-scheduling problem into a linear programming formulation. Consequently, the explosion operator, mutation operator, and selection operator to be designed are all tailored for permutation space. As shown in Table \ref{alp1}, experimental results indicate that under the gpt-4o-mini model, conventional automated heuristic design frameworks fail to adapt to the single-runway aircraft landing problem with complex constraints. Remarkably, Reevo and Funsearch independently arrived at the same heuristic rules through their search procedures. In contrast, our framework achieves near-optimal results in most tasks and even surpasses the reference optimum in certain instances.

\begin{table}[htbp]
\centering
\caption{Comparsions on Aircraft Landing Problems with gpt-4o-mini}
\begin{tabular}{lcccc}
\toprule
Problem Instances & EoH & ReEvo & Funsearch & Ours \\
\midrule
Airplane1 & 24.70\% & 57.85\% & 57.85\% & 100.00\% \\
Airplane2 & 32.70\% & 72.91\% & 72.91\% & 99.33\% \\
Airplane3 & 11.70\% & 28.57\% & 28.57\% & 100.00\% \\
Airplane4 & 44.90\% & 56.25\% & 56.25\% & 100.00\% \\
Airplane5 & 37.20\% & 43.54\% & 43.54\% & 100.00\% \\
Airplane6 & 100.00\% & 100.00\% & 100.00\% & 100.00\% \\
Airplane7 & 100.00\% & 39.00\% & 39.00\% & 100.00\% \\
Airplane8 & 0.00\% & 44.42\% & 44.42\% & 97.08\% \\
Airplane9 & 32.80\% & 55.02\% & 55.02\% & 119.63\% \\
Airplane10 & 44.80\% & 48.74\% & 48.74\% & 100.25\% \\
Airplane11 & 44.60\% & 70.13\% & 70.13\% & 142.42\% \\
\midrule
Average Performance & 43.04\% & 56.04\% & 56.04\% & 105.66\% \\
\bottomrule
\label{alp1}
\end{tabular}
\end{table}

\subsection{Equitable partition problem}
This task addresses the problem of partitioning a set of individuals, each characterized by multiple binary attributes. So the individuals can be represented as $\{x_i|1\leq i\leq n, x_i\in \{0,1\}^m\}$, into exactly 8 distinct groups. The objective is to achieve a balanced distribution of attribute values across these groups. Specifically, for each binary attribute, we aim to minimize the variation in the count of individuals possessing the attribute value '1' across different groups. The optimization criterion is defined as the sum, over all attributes, of the average absolute deviation between the count in each group and the mean count across all groups.

In the initial fireworks algorithm, our initialization process incorporates two distinct strategies. The first approach involves directly generating a sequence of numbers ranging from 1 to 8, while ensuring that all digits from 1 to 8 appear at least once. The second strategy employs a greedy assignment method, where each individual is allocated to a group while simultaneously minimizing the current absolute deviation measure.

As evidenced by the results in the table, our proposed method successfully resolves all problem instances. In contrast, traditional optimization algorithms such as Particle Swarm Optimization (PSO) and Differential Evolution (DE) demonstrate unsatisfactory performance on this problem. More importantly, the performance gap observed between the first five and the last five instances indicates that standard swarm intelligence and evolutionary algorithms struggle to adapt effectively to diverse data distributions. Among automated algorithm design frameworks, Funsearch delivers the most competitive performance. The final algorithm configuration discovered by Funsearch employs a greedy grouping initialization strategy coupled with a simulated annealing procedure but EoH and ReEvo do not employ any search strategy. While this configuration achieves perfect solutions on the first five instances, it fails to produce optimal results for instances epp6 through epp8.

\begin{table}[htbp]
\centering
\caption{Comparison on Equitable Partition Problem with gpt-4o-mini}
\label{tab:epp_comparison}
\begin{tabular}{lcccccc}
\toprule
Problems Instances & EoH & ReEvo & Funsearch & PSO & DE & Ours \\
\midrule
Epp1  & 100.00\% & 100.00\% & 100.00\% & 7.18\% & 3.91\% & 100.00\% \\
Epp2  & 100.00\% & 14.30\% & 100.00\% & 5.84\% & 4.05\% & 100.00\% \\
Epp3  & 100.00\% & 33.30\% & 100.00\% & 7.19\% & 4.08\% & 100.00\% \\
Epp4  & 100.00\% & 33.30\% & 100.00\% & 6.88\% & 3.81\% & 100.00\% \\
Epp5  & 100.00\% & 100.00\% & 100.00\% & 6.42\% & 3.74\% & 100.00\% \\
Epp6  & 90.20\% & 100.00\% & 88.50\% & 57.29\% & 43.15\% & 100.00\% \\
Epp7  & 81.00\% & 100.00\% & 89.50\% & 67.45\% & 42.89\% & 100.00\% \\
Epp8  & 75.30\% & 100.00\% & 94.80\% & 69.23\% & 47.07\% & 100.00\% \\
Epp9  & 84.10\% & 100.00\% & 100.00\% & 71.42\% & 46.69\% & 100.00\% \\
Epp10 & 92.90\% & 100.00\% & 100.00\% & 76.66\% & 56.81\% & 100.00\% \\
\midrule
\textbf{Average Performance} & 92.35\% & 78.09\% & 97.28\% & 37.56\% & 25.62\% & 100.00\% \\
\bottomrule
\end{tabular}
\end{table}

\subsection{Flow shop scheduling problem}
The flow shop scheduling problem addresses the challenge of sequencing a set of jobs through a series of machines in a fixed technological order. Given $n$ jobs and $m$ machines, the objective is to determine an optimal permutation of jobs that minimizes the makespan—the total completion time required to process all jobs through all machines. Each job must be processed on all machines in the same predetermined order, typically from machine 1 to machine $m$. The processing times are specified in an $n \times m$ matrix, where each element $p_{ij}$ represents the processing time of job $j$ on machine $i$. A solution is represented as a permutation of job indices defining the processing sequence. If the solution violates constraints (e.g., invalid job sequencing or machine order), it is deemed infeasible and receives no score.

Our search space comprises all permutations of integers from 1 to n, meaning each candidate solution represents a distinct permutation of the sequence 1 to n. The initial population is generated through a randomized initialization procedure.

As shown in Table \ref{fss}, for this specific task, while conventional algorithms such as DE and PSO achieve reasonably competent performance, those designed via LLM-driven evolution demonstrate superior effectiveness. It is noteworthy that the top-performing algorithms identified by frameworks such as EoH, ReEvo, and Funsearch all incorporate search mechanisms of varying complexity, moving beyond purely online constructive approaches. Notably, Funsearch hybridizes multiple components including the NEH heuristic, 2-opt local search, and simulated annealing. Our results achieve a performance level comparable to these state-of-the-art methods. 
\begin{table}[htbp]
\centering
\caption{Comparison on Flow Shop Scheduling Problem with gpt-4o-mini}
\label{tab:fssp_comparison}
\begin{tabular}{l c c c c c c}
\toprule
Problems Instances & EoH & ReEvo & Funsearch & PSO & DE & Ours \\
\midrule
Fss1  & 95.00\% & 95.00\% & 95.80\% & 94.39\% & 94.98\% & 95.55\% \\
Fss2  & 94.40\% & 94.90\% & 94.90\% & 93.94\% & 94.41\% & 94.74\% \\
Fss3  & 94.80\% & 99.30\% & 99.30\% & 95.52\% & 97.94\% & 99.07\% \\
Fss4  & 96.70\% & 98.10\% & 97.80\% & 96.12\% & 95.93\% & 97.67\% \\
Fss5  & 95.50\% & 96.80\% & 96.40\% & 93.03\% & 95.84\% & 96.42\% \\
Fss6  & 97.50\% & 97.50\% & 98.70\% & 94.70\% & 97.52\% & 98.13\% \\
Fss7  & 98.00\% & 98.00\% & 98.00\% & 97.02\% & 98.01\% & 98.18\% \\
Fss8  & 96.40\% & 96.60\% & 96.60\% & 94.67\% & 95.06\% & 96.97\% \\
Fss9  & 98.00\% & 98.00\% & 97.70\% & 93.95\% & 94.83\% & 98.05\% \\
Fss10 & 95.10\% & 97.70\% & 97.70\% & 94.63\% & 95.93\% & 97.65\% \\
\midrule
Average Performance & 96.14\% & 97.19 \% & 97.29\% & 94.80\% & 96.04\% & 97.24\% \\
\bottomrule
\label{fss}
\end{tabular}
\end{table}

\subsection{Uncapacitated p-median problem}
The uncapacitated $p$-median problem is a fundamental combinatorial optimization problem in facility location theory. Given a graph $G = (V, E)$ with $n$ vertices and a symmetric cost matrix $D \in \mathbb{R}^{n \times n}$ representing shortest-path distances between all vertex pairs, the objective is to select exactly $p$ vertices as medians (facilities) such that the sum of distances from each vertex to its nearest median is minimized.

In our initial firework algorithm, we employ a discrete encoding scheme where each individual is represented as a binary string of length n with exactly p ones. The same representation is adopted for both PSO and DE in our implementation.

As shown in Table \ref{pmu}, the methods based on LLM-automated algorithm evolution all demonstrate strong performance, significantly surpassing traditional swarm intelligence optimization algorithms. While Funsearch achieves the best average performance, our method is closely behind it.

\begin{table}[htbp]
\centering
\caption{Comparison on Uncapacitated p-Median Problem with gpt-4o-mini}
\label{tab:pmedian_comparison}
\begin{tabular}{l c c c c c c}
\toprule
Problem Instances & EoH & ReEvo & Funsearch & PSO & DE & Ours \\
\midrule
Pmu1  & 98.80\% & 100.00\% & 100.00\% & 96.74\% & 97.83\% & 100.00\% \\
Pmu2  & 99.40\% & 99.70\% & 99.70\% & 94.98\% & 94.42\% & 99.97\% \\
Pmu3  & 96.60\% & 100.00\% & 100.00\% & 95.01\% & 97.24\% & 100.00\% \\
Pmu4  & 98.30\% & 100.00\% & 100.00\% & 91.92\% & 92.55\% & 99.99\% \\
Pmu5  & 98.30\% & 99.60\% & 100.00\% & 86.58\% & 94.15\% & 99.90\% \\
Pmu6  & 97.50\% & 100.00\% & 100.00\% & 95.22\% & 95.73\% & 99.98\% \\
Pmu7  & 99.70\% & 100.00\% & 100.00\% & 92.69\% & 90.86\% & 99.99\% \\
Pmu8  & 99.40\% & 99.80\% & 99.80\% & 90.18\% & 85.91\% & 99.98\% \\
Pmu9  & 96.20\% & 99.80\% & 100.00\% & 86.84\% & 83.90\% & 99.58\% \\
Pmu10 & 96.90\% & 99.90\% & 99.80\% & 80.04\% & 84.15\% & 99.78\% \\
Pmu11 & 99.70\% & 100.00\% & 100.00\% & 95.07\% & 94.24\% & 100.00\% \\
Pmu12 & 99.70\% & 100.00\% & 99.90\% & 90.85\% & 88.43\% & 99.99\% \\
Pmu13 & 97.90\% & 99.80\% & 99.90\% & 89.66\% & 86.25\% & 99.87\% \\
Pmu14 & 98.50\% & 99.50\% & 100.00\% & 85.77\% & 81.89\% & 99.48\% \\
Pmu15 & 98.20\% & 99.50\% & 100.00\% & 79.96\% & 82.06\% & 99.71\% \\
Pmu16 & 99.10\% & 99.70\% & 99.90\% & 93.22\% & 92.86\% & 100.00\% \\
Pmu17 & 99.70\% & 99.80\% & 99.80\% & 89.08\% & 87.65\% & 99.79\% \\
Pmu18 & 98.70\% & 100.00\% & 99.70\% & 88.17\% & 84.19\% & 99.92\% \\
Pmu19 & 98.10\% & 99.50\% & 99.30\% & 84.86\% & 78.93\% & 99.61\% \\
\midrule
Average Performance & 98.46\% & 99.82\% & 99.88\% & 89.83\% & 89.12\% & 99.87\% \\
\bottomrule
\label{pmu}
\end{tabular}
\end{table}

\subsection{Further discussion}
\subsubsection{Prompts Analysis}
We examine the evolution of the prompt templates. The high-performance mutation prompts evolved under different large language models are presented in Appendix. The evolved templates significantly enhance the level of detail and technical specificity compared to the original, emphasizing performance-oriented constraints such as exploration-exploitation balance, convergence speed, and robustness. A common trend is the focus on targeted algorithm improvement and practicality, requiring unchanged code interfaces and the return of only complete code, reflecting the optimization of prompt precision and operability through automatic evolution.

The templates generated by different models vary in language style, level of detail, and constraint focus, demonstrating the impact of model capabilities on output quality. The template generated by \textbf{gpt-4o-mini} emphasizes conceptual innovation, with fluent language but relatively lenient requirements, encouraging creative improvements such as adaptive strategies and hybridization techniques, showing the model's strength in producing heuristic content, though it may lack deep technical specifics. The template from \textbf{Qwen3-32B} stresses theoretical foundations and engineering practices, requiring improvements based on "strong theoretical and practical foundations" and focusing on testability and reproducibility, indicating the model's robust ability to combine academic rigor with practicality, with outputs leaning towards methodological guidance. The template by \textbf{GPT-5} is extremely detailed, incorporating multiple technical constraints like the use of random number generators, complexity control, and reproducibility, with highly structured language resembling technical specifications, demonstrating the model's advanced logical reasoning and domain knowledge, capable of generating highly precise and secure prompts, though potentially at the cost of conciseness.

Different large language models require distinct guidance approaches, which in turn leads to their unique areas of emphasis. This inherent relationship validates the necessity of a co-evolutionary mechanism.

\begin{figure}
    \centering
    \includegraphics[width=1\linewidth]{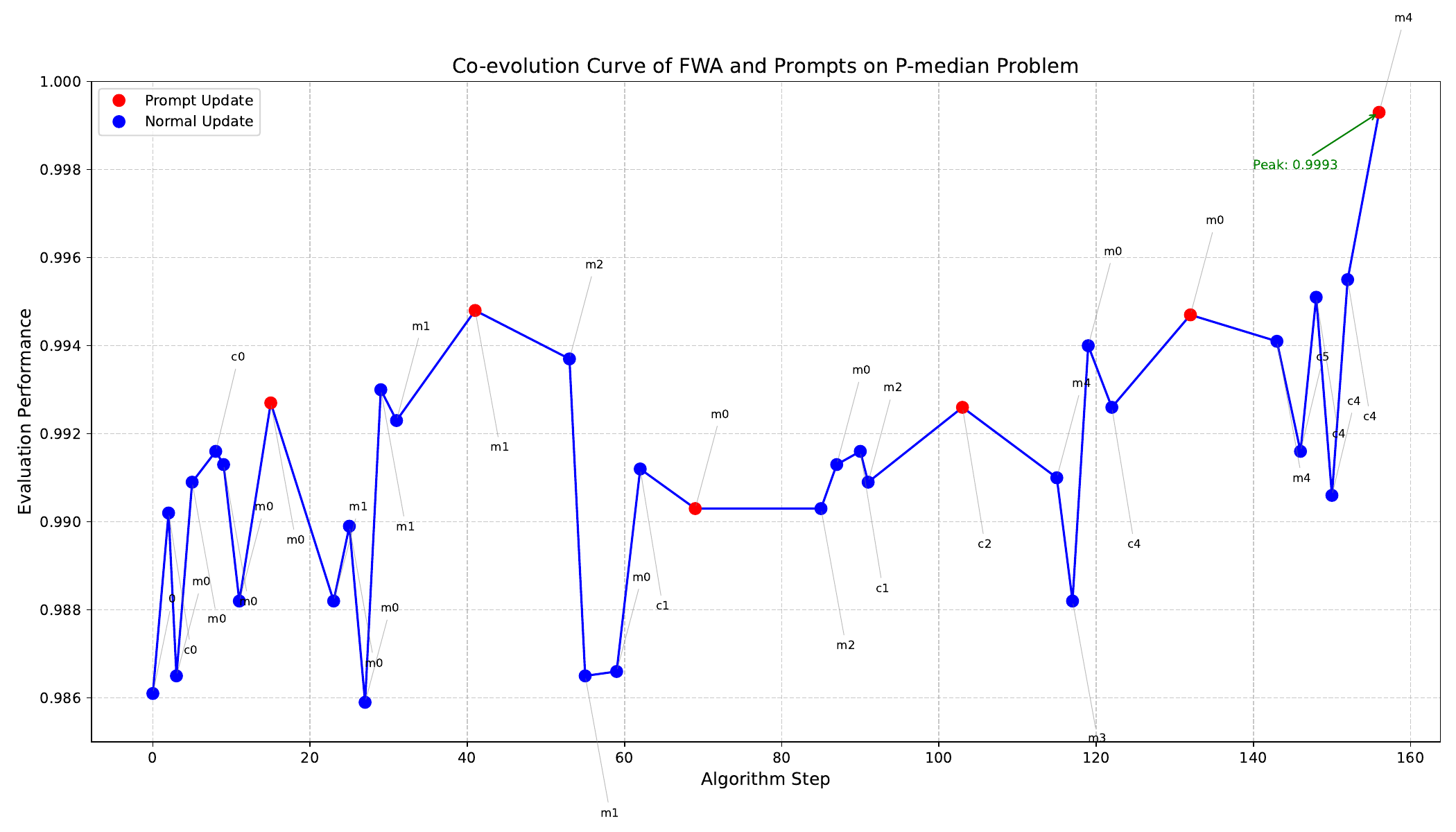}
    \caption{Evolution diagram of the complete LLM+FWA+P framework applied to one PMU problem instance, where red dots indicate the moments of prompt template updates (a stage that generates multiple algorithms), and blue dots represent the remaining evolutionary moments. We have annotated the best-performing prompt template at each moment, where $m_i$ denotes the i-th generation mutation prompt template, and $c_i$ denotes the i-th generation crossover prompt template. The diagram reveals that many local maxima are attained precisely at the moments when prompt templates are updated, with $m_4$ ultimately achieving the optimal performance.}
    \label{fig:evolution}
\end{figure}

Furthermore, the evolutionary trajectory of a single LLM+FWA+P run on the PMU task, illustrated in Figure \ref{fig:evolution}, demonstrates the critical importance of prompt template updates for enhancing overall performance.

\subsubsection{Ablation Study}
Next, we conduct a systematic ablation experiment to validate the efficacy of the LLM+FWA+P method. As shown in Table \ref{Abalation}, the full model achieves optimal performance on most problem instances, demonstrating particular superiority in the Airplane series (e.g., 142.42\% on Airplane11). The ablation study compares three progressively simplified methods: LLM+FWA+P, LLM+FWA, and initial FWA.

The results reveal a distinct performance hierarchy, confirming each component's necessity. The P component proves critical for complex problems, as its removal causes a 9.43\% performance drop on Airplane11. The LLM component's fundamental role is evidenced in Pmu tasks, where its absence leads to catastrophic performance degradation (e.g., from 99.78\% to 63.13\% on Pmu10).

Notably, the method maintains robust performance across tasks with different initialization strategies. It achieves significant optimization gains in Airplane/Epp tasks with enhanced initialization, while maintaining competitive advantage in Fss/Pmu tasks using direct encoding. This cross-task stability demonstrates that LLM+FWA+P constitutes a generalized solving framework whose effectiveness is independent of specific initialization techniques from human knowledge.

\begin{table}[htbp]
\centering
\caption{Performance comparison of different algorithms with gpt-4o-mini}
\label{tab:performance}
\begin{tabular}{lcccccc}
\toprule
\textbf{Problems} & \textbf{EoH} & \textbf{ReEvo} & \textbf{Funsearch} & \textbf{LLM + FWA + P} & \textbf{LLM + FWA} & \textbf{Initial FWA} \\
\midrule
Airplane2 & 32.70\% & 72.91\% & 72.91\% & 99.33\% & 99.20\% & 98.67\% \\
Airplane8 & 0.00\% & 44.42\% & 44.42\% & 97.08\% & 94.82\% & 78.63\% \\
Airplane9 & 32.80\% & 55.02\% & 55.02\% & 119.63\% & 114.93\% & 107.56\% \\
Airplane10 & 44.80\% & 48.74\% & 48.74\% & 100.25\% & 97.68\% & 88.00\% \\
Airplane11 & 44.60\% & 70.13\% & 70.13\% & 142.42\% & 132.99\% & 128.69\% \\
Epp7 & 81.00\% & 100.00\% & 89.50\% & 100.00\% & 100.00\% & 99.27\% \\
Epp8 & 75.30\% & 100.00\% & 94.80\% & 100.00\% & 99.02\% & 95.94\% \\
Fss1 & 95.00\% & 95.00\% & 95.80\% & 95.55\% & 95.41\% & 94.99\% \\
Fss2 & 94.40\% & 94.90\% & 94.90\% & 94.74\% & 94.58\% & 94.42\% \\
Pmu10 & 96.90\% & 99.90\% & 99.80\% & 99.78\% & 98.82\% & 63.13\% \\
Pmu15 & 98.20\% & 99.50\% & 100.00\% & 99.71\% & 98.34\% & 67.57\% \\
Pmu19 & 98.10\% & 99.50\% & 99.30\% & 99.92\% & 99.44\% & 73.24\% \\
\bottomrule
\label{Abalation}
\end{tabular}
\end{table}
\subsubsection{Impact of LLMs}
Meanwhile, we also compared the impact of different large language models (LLMs) on the final performance of various algorithm design frameworks. We selected three LLMs—gpt-4o-mini, Qwen3-32B, and gpt-5—and conducted comprehensive comparative experiments on the aircraft landing problem. The results are presented in Table \ref{different llms}， which clearly demonstrates the significant advantages of the LLM+FWA+P framework over comparative algorithms such as EoH, ReEvo, and Funsearch.  

First, LLM+FWA+P exhibits superior and consistent performance across different underlying LLMs. As shown in the "Average Performance" row, regardless of whether the base LLM is the relatively less powerful GPT-4o-mini or the highly capable GPT-5, the average performance of LLM+FWA+P (105.34\%, 109.36\%, 112.57\%) is significantly and consistently higher than all benchmarks. Notably, when using GPT-4o-mini, the performance of LLM+FWA+P is nearly double that of the next best algorithms (ReEvo and Funsearch, both at 56.04\%). This robustly validates the inherent effectiveness of the framework itself, which is capable of amplifying and complementing the capabilities of the underlying LLM, rather than being entirely dependent on the LLM's native strength.

Second, the LLM+FWA+P framework demonstrates breakthrough exploratory capabilities in solving complex problems. For more challenging instances such as Airplane9 to Airplane11, where the performance of other algorithms generally remains below 100\%, the LLM+FWA+P framework consistently achieves results significantly exceeding 100\% (e.g., 142.42\% for Airplane11 with GPT-4o-mini). This indicates that the framework not only finds feasible solutions but can surpass the established benchmark to discover superior solutions, highlighting its unique advantage in exploring the algorithm space.

Finally, the framework effectively leverages the potential of more powerful LLMs to enable further performance gains. When the underlying LLM is upgraded from GPT-4o-mini to GPT-5, the "Average Performance" of LLM+FWA+P increases from 105.34\% to 112.57\%, achieving peak performance close to 155\% on difficult problems (e.g., 154.34\% for Airplane11 with GPT-5). This suggests a positive synergistic effect between the LLM+FWA+P framework and advanced LLMs, where the framework successfully guides more capable LLMs to produce higher-quality algorithmic innovations.

\begin{table}[htbp]
\centering
\caption{Performance comparison of different algorithms with different LLMs on aircraft landing problems}
\label{tab:model_comparison}
\resizebox{\textwidth}{!}{%
\begin{tabular}{lcccccccccccc}
\toprule
& \multicolumn{4}{c}{\textbf{GPT-4o-mini}} & \multicolumn{4}{c}{\textbf{Qwen3-32B}} & \multicolumn{4}{c}{\textbf{GPT-5}} \\
\cmidrule(lr){2-5} \cmidrule(lr){6-9} \cmidrule(lr){10-13}
\textbf{Problems} & \textbf{EoH} & \textbf{ReEvo} & \textbf{Funsearch} & \textbf{LLM + FWA + P} & \textbf{EoH} & \textbf{ReEvo} & \textbf{Funsearch} & \textbf{LLM + FWA + P} & \textbf{EoH} & \textbf{ReEvo} & \textbf{Funsearch} & \textbf{LLM + FWA + P} \\
\midrule
Airplane1 & 24.70\% & 57.85\% & 57.85\% & 100.00\% & 57.90\% & 57.90\% & 100.00\% & 100.00\% & 100.00\% & 100.00\% & 100.00\% & 100.00\% \\
Airplane2 & 32.70\% & 72.91\% & 72.91\% & 99.33\% & 72.90\% & 72.90\% & 100.00\% & 100.00\% & 100.00\% & 100.00\% & 100.00\% & 100.00\% \\
Airplane3 & 11.70\% & 28.57\% & 28.57\% & 100.00\% & 28.60\% & 35.80\% & 100.00\% & 100.00\% & 100.00\% & 100.00\% & 100.00\% & 100.00\% \\
Airplane4 & 44.90\% & 56.25\% & 56.25\% & 100.00\% & 56.20\% & 56.20\% & 100.00\% & 100.00\% & 100.00\% & 100.00\% & 100.00\% & 100.00\% \\
Airplane5 & 37.20\% & 43.54\% & 43.54\% & 100.00\% & 43.50\% & 43.50\% & 100.00\% & 100.00\% & 100.00\% & 100.00\% & 100.00\% & 100.00\% \\
Airplane6 & 100.00\% & 100.00\% & 100.00\% & 100.00\% & 100.00\% & 100.00\% & 100.00\% & 100.00\% & 100.00\% & 100.00\% & 100.00\% & 100.00\% \\
Airplane7 & 100.00\% & 39.00\% & 39.00\% & 100.00\% & 39.00\% & 39.00\% & 100.00\% & 100.00\% & 100.00\% & 100.00\% & 100.00\% & 100.00\% \\
Airplane8 & 0.00\% & 44.42\% & 44.42\% & 97.08\% & 44.40\% & 53.90\% & 100.00\% & 100.00\% & 100.00\% & 98.70\% & 100.00\% & 100.00\% \\
Airplane9 & 32.80\% & 55.02\% & 55.02\% & 119.63\% & 55.00\% & 57.80\% & 71.60\% & 137.00\% & 137.50\% & 138.00\% & 138.30\% & 139.86\% \\
Airplane10 & 44.80\% & 48.74\% & 48.74\% & 100.25\% & 48.70\% & 58.50\% & 82.20\% & 116.04\% & 122.40\% & 135.30\% & 131.20\% & 144.06\% \\
Airplane11 & 44.60\% & 70.13\% & 70.13\% & 142.42\% & 70.10\% & 78.10\% & 73.00\% & 149.96\% & 143.10\% & 141.70\% & 152.30\% & 154.34\% \\
Average Performance & 43.04\% & 56.04\% & 56.04\% & 105.34\% & 56.03\% & 59.42\% & 93.35\% & 109.36\% & 109.36\% & 110.34\% & 111.07\% & 112.57\% \\
\bottomrule
\end{tabular}%
}
\label{different llms}
\end{table}

\subsubsection{Analysis on FWA generated by GPT-5}
To gain an intuitive and in-depth understanding of the algorithmic characteristics of the optimized FWA by GPT-5, we have included its complete code implementation in the Appendix and conducted a meticulous analysis focusing on its three core operators.

The explosion operator serves as the primary search mechanism, generating new spark solutions from each firework individual. Its core technology lies in the sequence weight construction mechanism, which calculates positional deviations between the current sequence and the target sequence, combined with aircraft penalty costs to form guided perturbation weights. The algorithm employs a multi-modal perturbation strategy with 50\% probability for swap operations, 30\% for insertion operations, and 20\% for reversal operations, each incorporating amplitude-based adaptive control. The number of explosion steps correlates positively with amplitude, ensuring dynamic search range adjustment according to fitness. Meanwhile, repair mechanisms and lightweight local optimization are introduced to fix infeasible solutions through adjacent swaps, and solution quality is enhanced through limited neighborhood searches. Finally, solution uniqueness is maintained via hash sets to avoid redundant computations.

The mutation operator functions as an auxiliary search mechanism, performing secondary optimization on sparks generated by explosion with emphasis on population diversity enhancement. This operator adopts a composite mutation operation system integrating four mutation operations with probabilistic distribution: swap operations (35\%), insertion operations (35\%), reversal operations (20\%), and or-opt block moving (10\%). The mutation process demonstrates a guidance-oriented strategy, with 60\% probability of selecting the aircraft with maximum positional deviation for priority mutation, and introduces classical Or-opt block moving operations to simulate professional domain search behaviors. An intelligent repair mechanism combines guided insertion and random swapping for hybrid repair of infeasible solutions, where guided insertion focuses on the aircraft with maximum deviation while random swapping increases exploration randomness. Mutation scale is controlled through mutation rate, targeting 20\% of total sparks for mutation, achieving balance between elitism preservation and diversity enhancement.

The selection operator employs a three-round progressive strategy combining elitism preservation and diversity maintenance to ensure continuous population quality improvement. The first round of elite selection preserves 60\% population size of optimal non-duplicate solutions, achieving strict deduplication based on sequence hashing. The second round of diversity selection introduces an adaptive distance threshold mechanism, using mean absolute position deviation as a diversity metric to filter individuals maintaining sufficient distance from selected solutions. The third round of supplemental selection fills remaining positions by fitness order, allowing moderate similarity. Sequence similarity measurement utilizes mean absolute difference of position mapping vectors, with thresholds adapting to problem scale. A fitness cleaning mechanism handles anomalies by setting non-finite values to infinity, ensuring numerical stability during selection. This layered selection strategy effectively prevents premature convergence and maintains population exploration capability.

\section{Conclusion}
This paper proposes a novel framework for LLM-driven swarm intelligence optimization algorithm design, whose core innovation lies in the introduction of a co-evolution mechanism for prompt templates. Departing fundamentally from prior automated algorithm design approaches (e.g., EoH, Funsearch, and Reevo) that focused predominantly on automating the algorithm structures themselves, our framework treats the prompt templates themselves as evolvable objects, enabling them to learn and adapt collaboratively with the algorithm population within the same evolutionary process. Through this design, the framework empowers the prompts to more effectively guide the contemporary large language models, thereby better unleashing their inherent algorithm design capabilities. Consequently, it significantly reduces reliance on the innate capabilities of the base LLM, rendering the entire algorithm evolution process more intelligent and comprehensive .

Extensive experiments on diverse NP-hard problem instances, supported by thorough ablation studies, demonstrate that our framework, by fully leveraging the guiding role of prompts, achieves superior performance across various optimization scenarios compared to several previous excellent benchmark frameworks, including EoH, Funsearch, and Reevo. These results not only validate the effectiveness of the prompt co-evolution mechanism but also mark a critical step forward for swarm intelligence optimization research paradigms towards greater efficiency and automation. We firmly believe that the research paradigm based on large language models, deeply integrated with prompt engineering, will inject new vitality into the field of swarm intelligence optimization and open up broader application prospects.
\bibliographystyle{plain} % 设定参考文献的显示样式
\bibliography{PromptFWA}     % 告诉LaTeX使用哪个.bib文件（不含后缀名）

\section{Appendix}
\subsection{Prompts produced by different LLMs}
\begin{lstlisting}[language=Python, caption={Good Mutation Prompt Generation Function by gpt-4o-mini}]
def get_mutation_prompts(problem_description, current_code, current_perfomance):
    introduction = f"You are an expert in algorithm design and combinatorial optimization. Your task is to improve a Firework Algorithm (FWA) for solving the following problem: {problem_description}.\n" \
                 + "You should focus on enhancing the algorithm's performance by redesigning one of its core components with creativity and rigor. Consider aspects such as exploration-exploitation balance, diversity preservation, convergence speed, and problem-specific characteristics. Think about how to incorporate adaptive strategies, hybridization with other techniques, or novel initialization methods if appropriate. The goal is to achieve better scalability, robustness, and solution quality."
    current_fwa = f"The current implementation of the FWA is as follows:\n{current_code}\nIts performance is currently {current_perfomance}. Higher values indicate better performance.\n" \
                 + "Analyze the code carefully and identify potential bottlenecks or areas where improvements can be made. Pay attention to how each component contributes to the overall behavior of the algorithm."
    requirements = "1. Choose exactly one of the functions among 'explode', 'mutate', or 'select' from the provided code and redesign it.\n" \
                 + "2. Ensure that your redesigned function maintains the same input and output interfaces as the original.\n" \
                 + "3. Do not modify any other parts of the algorithm.\n" \
                 + "4. Return only the complete updated FWA code in the format: <code>...</code>, with no additional explanations."
    prompt_mutation = introduction + current_fwa + requirements
    return prompt_mutation
\end{lstlisting}

\begin{lstlisting}[language=Python, caption={Good Mutation Prompt Generation Function by Qwen3-32B}]
def get_mutation_prompts(problem_description, current_code, current_performance):
    introduction = f"You are a world-class expert in algorithm engineering with deep expertise in evolutionary and metaheuristic optimization. Your task is to iteratively refine a fireworks algorithm (FWA) for the combinatorial optimization problem: {problem_description}. Each step should focus on a single meaningful and impactful improvement to one of the three core operators—'explode', 'mutate', or 'select'. Your objective is to enhance solution quality, convergence speed, and robustness while maintaining structural fidelity to the original algorithm. Use advanced algorithm design techniques, mathematical reasoning, and empirical insights to guide your changes. Your suggestions must be grounded in strong theoretical and practical foundations, and every modification should be performance-oriented, testable, and reproducible."
    current_fwa = f"The current FWA implementation, which achieves a performance score of {current_performance}, is provided below. You are to produce a refined version of the FWA by introducing one, and only one, focused change to one of the core operators. Preserve all function signatures, interfaces, and the overall algorithmic structure to ensure compatibility and continuity.\n\n" \
                  f"{current_code}"
    requirements = "1. Modify exactly one of the core operators: 'explode', 'mutate', or 'select'.\n" \
                  + "2. Do not change function signatures, control flow, or core algorithm structure.\n" \
                  + "3. Your modification must improve performance, robustness, or convergence speed in a measurable and meaningful way.\n" \
                  + "4. Return only the complete and syntactically correct updated FWA code, enclosed in <code>...</code>, with no explanations, comments, or additional text."
    prompt_mutation = introduction + "\n\n" + current_fwa + "\n\n" + requirements
    return prompt_mutation

\end{lstlisting} 

\begin{lstlisting}[language=Python, caption={Good Mutation Prompt Generation Function by GPT-5}]
def get_mutation_prompts(problem_description, current_code, current_perfomance):
    introduction = (
        f"You are an expert automatic algorithm designer specializing in combinatorial optimization and Fireworks Algorithm (FWA) "
        f"variants. Your job is to produce a single targeted redesign to maximize the given performance metric (higher is better). "
        f"Analyze the baseline internally and simulate the impact of your changes on exploration–exploitation balance, robustness, "
        f"and reproducibility, but output only code and no commentary in the final answer. The task context is: {problem_description}.\n"
    )
    current_fwa = (
        f"The baseline FWA implementation (treated as the exact ground truth) is provided as code:\n"
        f"{current_code}\n"
        f"Its current measured performance is: {current_perfomance}.\n"
        f"You must return the COMPLETE algorithm code, identical to the baseline except for ONE redesigned function chosen as specified below.\n"
    )
    requirements = (
        "Strict output and design requirements:\n"
        "1) Choose EXACTLY ONE function to redesign from this set: 'explode', 'mutate', or 'select'.\n"
        "2) Redesign ONLY the chosen function. Preserve its function name, parameter list (names, order, default values), input types, "
        "and return value(s) and their types/shapes. Do not modify any other functions, classes, global variables, helper utilities, "
        "control flow, or imports anywhere outside the chosen function.\n"
        "3) The modification must be drop-in compatible and runnable: the whole algorithm must execute without errors using the same inputs. "
        "Do not add parameters, change data structures, or alter public interfaces.\n"
        "4) Preserve reproducibility: reuse the baseline's existing random-number generator(s) and seed(s). Do NOT introduce new randomness sources, "
        "new global state, new external dependencies, or new imports. If the baseline uses a seeded RNG object, use it; do not switch to other RNGs.\n"
        "5) Aim to improve performance by balancing exploration and exploitation while keeping or reducing computational complexity relative to the original function. "
        "Examples of allowed strategies (within the chosen function only): adaptive amplitudes/step sizes based on fitness or diversity, diversity maintenance and "
        "duplicate handling, robust boundary handling (clip/reflect within known bounds), elite-aware adjustments. Avoid super-linear time/memory increases; keep "
        "overheads to constant or near-constant factors.\n"
        "6) Maintain coding style, variable names, typing hints, and IO contracts used in the baseline. Do not rename the chosen function or its parameters. "
        "Do not change return ordering, container types, or shapes. Handle edge cases robustly (e.g., ties, NaN/Inf fitness, empty lists, boundary-only candidates) "
        "without altering the external interface.\n"
        "7) Output formatting: RETURN ONLY the full algorithm code wrapped EXACTLY ONCE in <code>...</code> tags. Do not include explanations, comments, print/log statements, "
        "or any extra text before or after the tags. The returned code must be identical to the baseline except for the single redesigned function.\n"
        "8) Determinism: under the same seed, the algorithm must remain reproducible. If you change the order or number of random calls inside the chosen function, "
        "ensure the seeded RNG still governs all randomness. Do not introduce time-based or environment-based variability.\n"
        "9) Constraints checklist before returning: same imports, same module-level definitions, same public interfaces; no added dependencies; no I/O; no assertions that break runtime; "
        "no changes outside the chosen function. If uncertain, prefer minimal local changes that improve robustness and performance while honoring all constraints.\n"
    )
    prompt_mutation = introduction + current_fwa + requirements
    return prompt_mutation

\end{lstlisting} 

\subsection{Best FWA on Aircraft landing problem produced by GPT-5}
\begin{lstlisting}[language=Python, caption={Best code generated by LLM + FWA + P with GPT-5}]
import random
import numpy as np
from typing import List, Dict, Tuple
import copy
import time

class FWA:
    def __init__(
        self,
        evaluator: callable,  # Evaluation function: input -> fitness (smaller is better)
        num_planes: int,      # Number of planes
        num_runways: int,     # Number of runways
        planes: List[Dict],   # Plane information (contains "earliest"/"latest"/"target"/"penalty_early"/"penalty_late")
        separation: List[List[float]],  # Separation matrix (separation[i][j] = minimum separation when i precedes j)
        fw_size: int = 5,    # Number of fireworks (population size)
        sp_size: int = 20,   # Total number of sparks (total sparks generated by all fireworks)
        init_amp: float = 5,  # Initial explosion amplitude (time perturbation ratio relative to time window length)
        max_iter: int = 200,  # Maximum iterations
        mutation_rate: float = 0.2,  # Mutation probability
        lp_path, #Path of the lp function
    ):
        # Core parameter storage
        self.evaluator = evaluator
        self.num_planes = num_planes
        self.num_runways = num_runways
        self.planes = planes
        self.separation = np.array(separation)
        self.fw_size = fw_size
        self.sp_size = sp_size
        self.init_amps = init_amp
        self.max_iter = max_iter
        self.mutation_rate = mutation_rate
        self.silent = 0
        self.lp_path = lp_path 
        self.earliest_landing_times = np.array([self.planes[index]["earliest"] for index in range(self.num_planes)])
        self.target_landing_times = np.array([self.planes[index]["target"] for index in range(self.num_planes)])
        self.latest_landing_times = np.array([self.planes[index]["latest"] for index in range(self.num_planes)])
        self.penalty_early = np.array([self.planes[index]['penalty_early'] for index in range(self.num_planes)])
        self.penalty_late = np.array([self.planes[index]['penalty_late'] for index in range(self.num_planes)])
        # Dynamic parameters (adaptive explosion amplitude)
        self.amps = [init_amp] * fw_size  # Explosion amplitude for each firework
        self.population = []  # Population storage (tuple: (runway_assignment, landing_times))
    def sequence2time(self, sequence):
        # Here sequence is a numpy sorted array
        import sys
        sys.path.append(self.lp_path)
        from linprog import solve_sequence_with_cost
        n = self.num_planes
        a, b, c = self.earliest_landing_times[sequence], self.latest_landing_times[sequence], self.target_landing_times[sequence]
        d, e = self.penalty_early[sequence], self.penalty_late[sequence]
        s = self.separation[sequence][:,sequence]
        return solve_sequence_with_cost(n, a, b, c, d, e, s)

    def initialize(self):
        target_landing_times = [self.planes[index]["target"] for index in range(self.num_planes)]
        idx = np.argsort(target_landing_times)
        times, cost = self.sequence2time(idx)
        runway_assignment = [1 for i in range(self.num_planes)]
        final_times = [times[idx.tolist().index(k)] for k in range(self.num_planes)]
        return runway_assignment, final_times, idx
    
    def explode(self, firework, amp):
        runway_assignment, times, sequence = firework
        n = self.num_planes
        sp_per_fw = max(1, self.sp_size // self.fw_size)
        exploded = []

        # Target order (sorted by target time)
        target_order = np.argsort(self.target_landing_times)
        target_pos = np.empty(n, dtype=int)
        target_pos[target_order] = np.arange(n)

        # Inertia/guidance: weighted point selection based on deviation from target sequence and penalty strength
        penalties = self.penalty_early + self.penalty_late

        def build_weights(idx_arr):
            pos = np.empty(n, dtype=int)
            pos[idx_arr] = np.arange(n)
            displacement = np.abs(pos - target_pos)
            # Avoid all zeros, add baseline of 1
            w = 1.0 + displacement * (1.0 + penalties)
            w = w.astype(float)
            s = w.sum()
            if s <= 0:
                w = np.ones_like(w, dtype=float) / n
            else:
                w = w / s
            return w, pos

        def apply_insertion(idx_arr, pos_map, k_plane, new_pos):
            cur_pos = pos_map[k_plane]
            if new_pos == cur_pos:
                return idx_arr
            # Perform insertion
            if new_pos < cur_pos:
                # Move left
                new_idx = np.concatenate([idx_arr[:new_pos], [k_plane], idx_arr[new_pos:cur_pos], idx_arr[cur_pos+1:]])
            else:
                # Move right
                new_idx = np.concatenate([idx_arr[:cur_pos], idx_arr[cur_pos+1:new_pos+1], [k_plane], idx_arr[new_pos+1:]])
            return new_idx

        def local_improve(idx_arr, base_cost, max_trials=3):
            # Lightweight local improvement: try several adjacent swaps, accept if LP cost decreases
            best_idx = idx_arr
            best_times, best_cost = self.sequence2time(best_idx)
            if not np.isfinite(best_cost):
                return None, None, np.inf
            trials = max_trials
            while trials > 0:
                trials -= 1
                p = random.randrange(n - 1)
                cand = best_idx.copy()
                cand[p], cand[p+1] = cand[p+1], cand[p]
                t, c = self.sequence2time(cand)
                if np.isfinite(c) and c < best_cost:
                    best_idx, best_times, best_cost = cand, t, c
            return best_idx, best_times, best_cost

        seen = set()
        seen.add(tuple(sequence.tolist()))

        # Each spark attempts to generate sp_per_fw unique candidates
        for _ in range(sp_per_fw):
            tries = 0
            generated = False
            while tries < 10 and not generated:
                tries += 1
                idx = sequence.copy()
                w, pos_map = build_weights(idx)

                # Perturbation steps related to amplitude
                steps = max(1, int(np.random.randint(1, int(amp) + 1)))

                for _step in range(steps):
                    move_type = np.random.choice([0, 1, 2], p=[0.5, 0.3, 0.2])  # 0:swap, 1:insertion, 2:reverse
                    if move_type == 0:
                        # Weighted selection of two different planes for swapping
                        i_plane = np.random.choice(n, p=w)
                        j_plane = np.random.choice(n, p=w)
                        # If duplicate, randomly pick different one
                        if i_plane == j_plane:
                            j_plane = (j_plane + 1) % n
                        i_pos = pos_map[i_plane]
                        j_pos = pos_map[j_plane]
                        idx[i_pos], idx[j_pos] = idx[j_pos], idx[i_pos]
                    elif move_type == 1:
                        # Insertion: move high-weight plane toward target position
                        k_plane = np.random.choice(n, p=w)
                        tgt = target_pos[k_plane]
                        # New position around target with some noise (larger amplitude, wider range)
                        jitter = int(np.random.normal(0, max(1, amp // 2)))
                        new_pos = int(np.clip(tgt + jitter, 0, n - 1))
                        # Need latest pos_map
                        _, pos_map = build_weights(idx)
                        idx = apply_insertion(idx, pos_map, k_plane, new_pos)
                    else:
                        # Reverse a segment
                        length = max(2, min(n, int(np.random.randint(2, min(n, int(amp) + 2)))))
                        start = np.random.randint(0, n - length + 1)
                        end = start + length
                        idx[start:end] = idx[start:end][::-1]

                key = tuple(idx.tolist())
                if key in seen:
                    continue

                times_cand, cost_cand = self.sequence2time(idx)

                # Repair attempt: if infeasible, perform minor adjacent swap repair
                if not np.isfinite(cost_cand):
                    repair_attempts = 3
                    repaired = False
                    cand = idx
                    for _r in range(repair_attempts):
                        p = random.randrange(n - 1)
                        cand2 = cand.copy()
                        cand2[p], cand2[p+1] = cand2[p+1], cand2[p]
                        t2, c2 = self.sequence2time(cand2)
                        if np.isfinite(c2):
                            idx = cand2
                            times_cand, cost_cand = t2, c2
                            repaired = True
                            break
                        else:
                            cand = cand2
                    if not repaired:
                        continue

                # Local improvement (small scale to control time)
                idx_impr, times_impr, cost_impr = local_improve(idx, cost_cand, max_trials=min(3, int(amp)))
                if idx_impr is not None and np.isfinite(cost_impr):
                    idx, times_cand, cost_cand = idx_impr, times_impr, cost_impr

                # Final feasibility check
                if np.isfinite(cost_cand):
                    # Construct final times (in plane ID order)
                    final_times = np.zeros(n, dtype=float)
                    for pos, pid in enumerate(idx):
                        final_times[pid] = times_cand[pos]
                    exploded.append(([1 for _ in range(n)], final_times, idx))
                    seen.add(tuple(idx.tolist()))
                    generated = True

        return exploded

    def mutate(self, sparks: List[Tuple[List[int], List[float]]]) -> List[Tuple[List[int], List[float]]]:
        """Mutation operator: additional perturbation on sparks"""
        n = self.num_planes
        mutated: List[Tuple[List[int], List[float]]] = []
        if not sparks:
            return mutated

        # Target order and weight preparation
        target_order = np.argsort(self.target_landing_times)
        target_pos = np.empty(n, dtype=int)
        target_pos[target_order] = np.arange(n)
        penalties = (self.penalty_early + self.penalty_late).astype(float)

        def build_weights(idx_arr: np.ndarray):
            pos_map = np.empty(n, dtype=int)
            pos_map[idx_arr] = np.arange(n)
            displacement = np.abs(pos_map - target_pos)
            w = 1.0 + (displacement + 1.0) * (penalties + 1.0)
            s = float(np.sum(w))
            if s <= 0:
                return np.ones(n) / n, pos_map, displacement
            return (w / s), pos_map, displacement

        def apply_insertion(idx_arr: np.ndarray, pos_map: np.ndarray, k_plane: int, new_pos: int):
            cur_pos = pos_map[k_plane]
            if new_pos == cur_pos:
                return idx_arr
            if new_pos < cur_pos:
                new_idx = np.concatenate([idx_arr[:new_pos], [k_plane], idx_arr[new_pos:cur_pos], idx_arr[cur_pos+1:]])
            else:
                new_idx = np.concatenate([idx_arr[:cur_pos], idx_arr[cur_pos+1:new_pos+1], [k_plane], idx_arr[new_pos+1:]])
            return new_idx

        def local_improve(idx_arr: np.ndarray, max_trials: int = 4):
            best_idx = idx_arr
            best_times, best_cost = self.sequence2time(best_idx)
            if not np.isfinite(best_cost):
                return None, None, np.inf
            trials = max_trials
            while trials > 0:
                trials -= 1
                # Two neighborhood types: adjacent swap or small segment reversal
                if random.random() < 0.6 and n >= 2:
                    p = random.randrange(n - 1)
                    cand = best_idx.copy()
                    cand[p], cand[p+1] = cand[p+1], cand[p]
                else:
                    if n >= 3:
                        L = random.randrange(2, min(8, n) + 1)
                        start = random.randrange(0, n - L + 1)
                        end = start + L
                        cand = best_idx.copy()
                        cand[start:end] = cand[start:end][::-1]
                    else:
                        continue
                t, c = self.sequence2time(cand)
                if np.isfinite(c) and c < best_cost:
                    best_idx, best_times, best_cost = cand, t, c
            return best_idx, best_times, best_cost

        def mutate_once(base_idx: np.ndarray):
            # Operation selection
            w, pos_map, disp = build_weights(base_idx)
            idx = base_idx.copy()
            # Compound mutation steps (1-2 steps)
            steps = 1 + int(random.random() < 0.5)
            for _ in range(steps):
                op = np.random.choice([0, 1, 2, 3], p=[0.35, 0.35, 0.2, 0.1])  # swap, insertion, reverse, or-opt
                if op == 0 and n >= 2:
                    # Weighted swap
                    i_plane = np.random.choice(n, p=w)
                    j_plane = np.random.choice(n, p=w)
                    if i_plane == j_plane:
                        j_plane = (j_plane + 1) % n
                    i_pos, j_pos = pos_map[i_plane], pos_map[j_plane]
                    idx[i_pos], idx[j_pos] = idx[j_pos], idx[i_pos]
                    # Update mapping
                    pos_map[idx[i_pos]] = i_pos
                    pos_map[idx[j_pos]] = j_pos
                elif op == 1:
                    # Insertion toward target
                    # Prefer planes with larger deviation
                    if random.random() < 0.6:
                        k_plane = int(np.argmax(disp))
                    else:
                        k_plane = np.random.choice(n, p=w)
                    tgt = target_pos[k_plane]
                    sigma = max(1, n // 10)
                    new_pos = int(np.clip(tgt + int(np.random.normal(0, sigma)), 0, n - 1))
                    # Need latest pos_map
                    w, pos_map, disp = build_weights(idx)
                    idx = apply_insertion(idx, pos_map, k_plane, new_pos)
                    # Recompute mapping
                    w, pos_map, disp = build_weights(idx)
                elif op == 2 and n >= 3:
                    # Segment reversal
                    L = random.randrange(2, min(8, n) + 1)
                    start = random.randrange(0, n - L + 1)
                    end = start + L
                    idx[start:end] = idx[start:end][::-1]
                    w, pos_map, disp = build_weights(idx)
                elif op == 3:
                    # Or-opt small block move
                    l = random.choice([1, 2, 3]) if n >= 4 else 1
                    if n > l:
                        start = random.randrange(0, n - l + 1)
                        block = idx[start:start + l].copy()
                        rest = np.concatenate([idx[:start], idx[start + l:]])
                        # Target position around mean target of block
                        tgt_block = int(np.round(np.mean(target_pos[block])))
                        jitter = int(np.random.normal(0, max(1, n // 12)))
                        ins_pos = int(np.clip(tgt_block + jitter, 0, n - l))
                        new_idx = np.concatenate([rest[:ins_pos], block, rest[ins_pos:]])
                        idx = new_idx
                        w, pos_map, disp = build_weights(idx)
            return idx

        # Construct deduplication set (including original sparks)
        seen = set()
        for spark in sparks:
            idx = spark[2]
            if isinstance(idx, np.ndarray):
                seen.add(tuple(idx.tolist()))
            else:
                seen.add(tuple(list(idx)))

        # Target mutation count
        target_count = max(1, int(len(sparks) * self.mutation_rate))

        # Randomly select spark indices to mutate
        indices = list(range(len(sparks)))
        random.shuffle(indices)
        indices = indices[:min(target_count, len(sparks))]

        for k in indices:
            _, _, base_idx = sparks[k]
            if not isinstance(base_idx, np.ndarray):
                base_idx = np.array(base_idx, dtype=int)
            attempts = 6
            success = False
            best_pack = None  # (idx, times, cost)
            while attempts > 0 and not success:
                attempts -= 1
                cand_idx = mutate_once(base_idx)
                key = tuple(cand_idx.tolist())
                if key in seen:
                    continue
                times_cand, cost_cand = self.sequence2time(cand_idx)
                if not np.isfinite(cost_cand):
                    # Repair: combine adjacent swap and guided insertion
                    repaired = False
                    repair_tries = 5
                    cur = cand_idx.copy()
                    for _r in range(repair_tries):
                        # Perform guided insertion on plane with large deviation
                        w0, pos0, disp0 = build_weights(cur)
                        plane_focus = int(np.argmax(disp0)) if random.random() < 0.7 else np.random.choice(n, p=w0)
                        tgt = target_pos[plane_focus]
                        new_pos = int(np.clip(tgt + int(np.random.normal(0, max(1, n // 12))), 0, n - 1))
                        cur = apply_insertion(cur, pos0, plane_focus, new_pos)
                        # Add random adjacent swap
                        if n >= 2 and random.random() < 0.6:
                            p = random.randrange(n - 1)
                            cur[p], cur[p+1] = cur[p+1], cur[p]
                        t2, c2 = self.sequence2time(cur)
                        if np.isfinite(c2):
                            cand_idx, times_cand, cost_cand = cur, t2, c2
                            repaired = True
                            break
                    if not repaired:
                        continue
                # Small-scale local improvement
                imp_idx, imp_times, imp_cost = local_improve(cand_idx, max_trials=4)
                if imp_idx is not None and np.isfinite(imp_cost):
                    cand_idx, times_cand, cost_cand = imp_idx, imp_times, imp_cost
                # Record
                best_pack = (cand_idx, times_cand, cost_cand)
                success = True

            if success and best_pack is not None and np.isfinite(best_pack[2]):
                idx_final, times_final, _ = best_pack
                final_times = np.zeros(n, dtype=float)
                for pos, pid in enumerate(idx_final):
                    final_times[pid] = times_final[pos]
                mutated.append(([1 for _ in range(n)], final_times, idx_final))
                seen.add(tuple(idx_final.tolist()))

            # Control size
            if len(mutated) >= target_count:
                break

        return mutated

    def select(self, population: List[Tuple[List[int], List[float]]], 
               sparks: List[Tuple[List[int], List[float]]], 
               mutated_sparks: List[Tuple[List[int], List[float]]]) -> List[Tuple[List[int], List[float]]]:
        """Selection operator: combine candidate solutions and select optimal population"""
        # Combine all candidate solutions
        candidates = population + sparks + mutated_sparks
        if not candidates:
            return population

        # Evaluate fitness (call user-provided evaluator)
        fitness = [self.evaluator.compute((ind[0], ind[1])) for ind in candidates]
        # Handle infeasible or abnormal values
        cleaned_fitness = []
        for f in fitness:
            if f is None or (isinstance(f, float) and not np.isfinite(f)) or (isinstance(f, np.ndarray) and not np.isfinite(float(f))):
                cleaned_fitness.append(float('inf'))
            else:
                try:
                    cleaned_fitness.append(float(f))
                except Exception:
                    cleaned_fitness.append(float('inf'))

        # Stable sorting (by fitness, index as tie-breaker for determinism)
        sorted_indices = sorted(range(len(candidates)), key=lambda i: (cleaned_fitness[i], i))

        # Precompute sequence inverse mapping for distance measurement
        n = self.num_planes
        seq_keys = []
        pos_maps = []
        for i in range(len(candidates)):
            idx = candidates[i][2]
            if isinstance(idx, np.ndarray):
                arr = idx
            else:
                arr = np.array(idx, dtype=int)
            seq_keys.append(tuple(arr.tolist()))
            pm = np.empty(n, dtype=int)
            pm[arr] = np.arange(n)
            pos_maps.append(pm)

        # Diversity threshold (based on average position deviation)
        diversity_threshold = max(1.0, float(n) / 12.0)

        selected_idx: List[int] = []
        seen_seq = set()

        # Elite proportion: prioritize selecting several optimal solutions (avoid premature discarding of good solutions)
        elite_quota = max(1, int(self.fw_size * 0.6))

        # First round: select elites (non-duplicate sequences)
        for i in sorted_indices:
            key = seq_keys[i]
            if key in seen_seq:
                continue
            selected_idx.append(i)
            seen_seq.add(key)
            if len(selected_idx) >= elite_quota:
                break

        # Second round: select remaining individuals based on diversity (avoid too similar to already selected)
        if len(selected_idx) < self.fw_size:
            for i in sorted_indices:
                if len(selected_idx) >= self.fw_size:
                    break
                if i in selected_idx:
                    continue
                key = seq_keys[i]
                if key in seen_seq:
                    continue
                # Calculate minimum distance to selected set
                min_dist = float('inf')
                pm_i = pos_maps[i]
                for j in selected_idx:
                    pm_j = pos_maps[j]
                    # Average absolute position difference
                    dist = float(np.mean(np.abs(pm_i - pm_j)))
                    if dist < min_dist:
                        min_dist = dist
                        if min_dist <= diversity_threshold:  # Early stopping
                            break
                if min_dist >= diversity_threshold:
                    selected_idx.append(i)
                    seen_seq.add(key)

        # Third round: if still insufficient, fill by fitness (allow somewhat similar but non-duplicate sequences)
        if len(selected_idx) < self.fw_size:
            for i in sorted_indices:
                if len(selected_idx) >= self.fw_size:
                    break
                if i in selected_idx:
                    continue
                key = seq_keys[i]
                if key in seen_seq:
                    continue
                selected_idx.append(i)
                seen_seq.add(key)

        # Fallback: if still insufficient (extreme case), allow duplicate sequences to ensure size
        if len(selected_idx) < self.fw_size:
            for i in sorted_indices:
                if len(selected_idx) >= self.fw_size:
                    break
                if i not in selected_idx:
                    selected_idx.append(i)

        return [candidates[i] for i in selected_idx[:self.fw_size]]

    def optimize(self) -> Tuple[List[int], List[float], float]:
        """Main optimization process"""
        # Initialize population (ensure all individuals are valid)
        self.population = []
        while len(self.population) < self.fw_size:
            runway_assignment, times, idx = self.initialize()
            self.population.append((runway_assignment, times, idx))
        
        best_fitness = float('inf')
        best_individual = None
        start_time = time.time()
        while not self.evaluator.stop():
            fitness = [self.evaluator.compute((ind[0], ind[1])) for ind in self.population]
            current_best_idx = np.argmin(fitness)
            current_best_fit = fitness[current_best_idx]
            current_best_ind = self.population[current_best_idx]
            
            # Update global best
            if current_best_fit < best_fitness:
                best_fitness = current_best_fit
                best_individual = current_best_ind
                self.silent = 0
            else:
                self.silent += 1
            
            # Adaptive adjustment of explosion amplitude (better fitness, larger amplitude)
            max_fit = max(fitness)
            self.amps = [
                max(1, min(int(self.init_amps * (1.5 - f/max_fit)), self.fw_size))
                for f in fitness
            ]
            # Generate explosion sparks
            all_sparks = []
            for i in range(self.fw_size):
                sparks = self.explode(self.population[i], self.amps[i])
                all_sparks.extend(sparks)
            
            # Mutation processing
            mutated_sparks = self.mutate(all_sparks)
            
            # Select next generation
            self.population = self.select(self.population, all_sparks, mutated_sparks)
            if self.silent >= 10:
                break
        return self.evaluator.get_best_solution(), self.evaluator.get_best_fitness()
\end{lstlisting}

\end{document}